\documentclass[10pt,twocolumn,letterpaper]{article}
\usepackage[OT1]{fontenc}
\usepackage[pagenumbers]{cvpr}
\usepackage{amsmath}
\usepackage{amssymb}
\usepackage{booktabs}
\usepackage{multirow}
\usepackage{float}
\usepackage{placeins}
\usepackage{graphicx}
\usepackage{tikz}
\usetikzlibrary{arrows.meta,calc,fit,positioning}
\usepackage{xcolor}
\usepackage{xspace}
\usepackage[table]{xcolor}
\definecolor{radarAccuracy}{RGB}{210,78,78}
\definecolor{radarSpeed}{RGB}{70,96,170}
\definecolor{radarAccuracyEdge}{RGB}{170,38,48}
\definecolor{radarSpeedEdge}{RGB}{42,58,128}
\definecolor{radarTokens}{RGB}{73,145,118}
\definecolor{radarTokensEdge}{RGB}{32,108,81}
\definecolor{radarGroup}{RGB}{238,174,105}

\definecolor{riseBlue}{RGB}{62,105,176}
\definecolor{riseViolet}{RGB}{118,91,171}
\definecolor{riseGreen}{RGB}{47,137,109}
\definecolor{riseOrange}{RGB}{216,137,66}
\definecolor{riseInk}{RGB}{50,58,72}

\definecolor{cvprblue}{rgb}{0.21,0.49,0.74}
\usepackage[pagebackref,breaklinks,colorlinks,allcolors=cvprblue]{hyperref}
\def\confName{CVPR}
\def\confYear{2027}
\title{GUI-PRISM: Efficient Action Decoding for GUI Agents}
\hypersetup{pdftitle={GUI-PRISM: Efficient Action Decoding for GUI Agents},pdfauthor={Anonymous manuscript}}

\author{
Enqi Liu$^{1,2}$, 
Zhi Gao$^{1,2}$\footnotemark[1], 
Liu Liu$^{4}$,
Lifeng Fan$^{2}$, 
Liyuan Pan$^{1,3}$\thanks{Corresponding author.}, \\
$^{1}$ Beijing Institute of Technology, Beijing, China\\
$^{2}$ Beijing Institute for General Artificial Intelligence, Beijing, China\\
$^{3}$ Yangtze Delta Region Academy of Beijing Institute of Technology, Jiaxing, China \\
$^{4}$ Huawei Technologies Ltd, Beijing, China\\
{\tt\small \{enqi.liu, zhi.gao, liyuan.pan\}@bit.edu, liuliu33@huawei.com, lifengfan@bigai.ai}
}

\usepackage{needspace}
\begin{document}
\raggedbottom
\maketitle
\suppressfloats[t]

\begin{abstract}
Graphical user interface (GUI) agents translate screenshots and user instructions into executable actions.
Most existing approaches rely on vision-language models to generate actions as text, often leading to substantial inference latency due to autoregressive decoding. To address this inefficiency, we propose GUI-PRISM, a hybrid action decoder that directly predicts action types and structured arguments while retaining language generation for open-ended text. Unlike existing spatial decoding methods that primarily focus on target localization, GUI-PRISM extends direct prediction to action types and their associated arguments, enabling complete GUI action prediction. Specifically, GUI-PRISM employs field-specific queries to extract action-relevant semantic and spatial information from a shared multimodal feature bank. It further models dependencies among action components through conditional queries. Gated residual cross-attention refines these predictions using fine-grained multimodal information. GUI-PRISM achieves 95.1\% accuracy on ScreenSpot-v2, outperforming the strongest head-based spatial decoding method by 2.0 percentage points while delivering a 3.8$\times$ inference speedup over the baseline. On AndroidControl-v2 and MM-Mind2Web-v2, it improves complete-action accuracy by 4.8 and 0.7 percentage points while achieving 2.7$\times$ and 3.9$\times$ inference speedups over the baseline, respectively. Our code is available on \href{https://github.com/Liu-arch/GUI-PRISM}{our project page}.
\end{abstract}

\begin{figure}[t]
\centering
\includegraphics[width=\columnwidth]{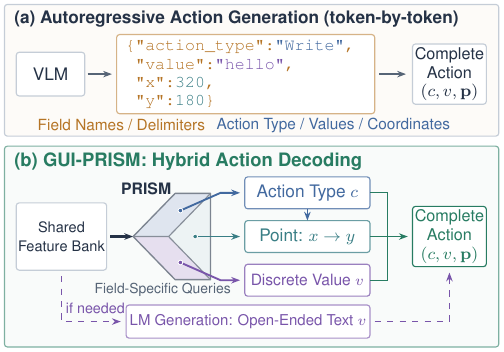}
\caption{\textbf{Comparison of GUI action decoding approaches.} (a) Autoregressive action generation sequentially decodes both schema formatting tokens (e.g., `\texttt{action\_type}' and `\texttt{:}') and action content (e.g., `\texttt{Write}', `\texttt{hello}', and `\texttt{320, 180}'). (b) GUI-PRISM directly predicts action types and structured arguments while retaining language generation for open-ended text.}
\label{fig:concept}
\end{figure}
\section{Introduction}
Graphical user interface (GUI) agents enable automated interaction with software applications by translating visual observations and user instructions into executable actions~\cite{xie2024osworld,wang2025opencua}. Accurate and efficient GUI interaction requires selecting appropriate action types and predicting their associated arguments, such as target coordinates and text inputs. Recent advances in vision-language models (VLMs) have substantially improved GUI action prediction, enabling agents to interpret diverse instructions, identify interaction targets, and generate complete actions across different interfaces~\cite{lin2025showui,qin2025uitars,yang2026guilibra}.

Despite these advances, generating GUI actions as text introduces substantial autoregressive decoding overhead, as illustrated in Fig.~\ref{fig:concept}(a). Models must sequentially generate both schema formatting tokens, such as field names `\texttt{action\_type, value, x, y}' and delimiters such as `\texttt{", :, \{, \}}', and action content, including action type `\texttt{Write}', the text value `\texttt{hello}', and coordinate values `\texttt{320, 180}'. These additional decoding steps increase inference latency before actions can be executed.

To reduce this decoding overhead, we explore using a unified action head to directly predict action types and structured arguments, avoiding their token-by-token generation. Recent spatial decoding methods primarily focus on improving target localization through specialized visual representations and spatial decoding mechanisms~\cite{wu2025guiactor,chen2026v2p,jinggrounding}. However, complete GUI action prediction requires handling a broader range of outputs, including action types, target coordinates, discrete arguments, and open-ended text. 
This motivates a hybrid decoding approach that combines direct prediction of structured action components with language generation for open-ended text.

Realizing this approach presents two key challenges.
(1) Extracting field-specific multimodal information:
Different prediction heads require distinct information from the same screenshot and instruction. For example, clicking the `\texttt{Save}' button requires determining the intended action and precisely locating its target. 
These differing information requirements make it difficult for each prediction head to selectively extract the action-relevant semantic or spatial information it needs from shared multimodal representations.
(2) Preserving dependencies in direct action decoding:
Unlike autoregressive generation, which naturally conditions subsequent tokens on preceding outputs, direct prediction does not inherently capture dependencies among action components. The predicted action type determines the required arguments, while coordinate prediction involves dependencies between the horizontal and vertical axes. The challenge is to preserve these relationships without reverting to token-by-token generation.

\suppressfloats[t]
\begin{table}[t]
\centering
\caption{GUI-PRISM action types, required parameters, and illustrative examples. Open-ended text values use the LM; finite values use classifiers. \texttt{None} denotes the native empty-value label.}
\label{tab:action-space}
\footnotesize
\setlength{\tabcolsep}{2pt}
\renewcommand{\arraystretch}{1.05}
\begin{tabular*}{\columnwidth}{@{\extracolsep{\fill}}llll@{}}
\toprule
Operation $c$ & Coordinate $\mathbf p$ & Value $v$ & Example \\
\midrule
Click         & Required & Finite label   & Click Submit \\
LongPress     & Required & \texttt{None}  & Hold an icon \\
Write         & Optional & Text           & Enter ``hello'' \\
Select        & Required & Text           & Select ``English'' \\
Scroll        & Optional & Direction      & Scroll down \\
Swipe         & --       & Direction      & Swipe left \\
KeyboardPress & Optional & Key            & Press Enter \\
OpenApp       & --       & Text           & Open Settings \\
NavigateBack  & --       & \texttt{None}  & Go back \\
NavigateHome  & --       & \texttt{None}  & Go home \\
Wait          & --       & Duration       & Wait 2 seconds \\
Terminate     & --       & Status         & Mark completion \\
Answer        & --       & Text           & Reply ``Done'' \\
\bottomrule
\end{tabular*}
\end{table}

In this paper, we propose GUI-PRISM, a hybrid action decoder that directly predicts action types and structured arguments while retaining language generation for open-ended text values when required, as illustrated in Fig.~\ref{fig:concept}(b).  
Specifically, GUI-PRISM introduces field-specific queries to extract action-relevant semantic and spatial information from a shared PRISM feature bank integrating visual, contextual, and action-relevant features.
We explicitly model dependencies among action components: the predicted action type determines the applicable arguments and their corresponding predictors, while coordinate queries incorporate the selected action type and the initial horizontal-coordinate estimate.
We further employ gated residual cross-attention to refine these predictions using fine-grained action-relevant and visual information from the shared PRISM feature bank.
Together, these designs enable GUI-PRISM to predict complete GUI actions while preserving dependencies among structured action components.

GUI-PRISM achieves 95.1\% accuracy on ScreenSpot-v2, outperforming the strongest head-based spatial decoding method by 2.0 percentage points while delivering a 3.8$\times$ inference speedup over the baseline. 
On AndroidControl-v2 and MM-Mind2Web-v2, it improves complete-action accuracy by 4.8 and 0.7 percentage points over the baseline while achieving 2.7$\times$ and 3.9$\times$ inference speedups, respectively.

Our main contributions are as follows:
\begin{itemize}
\setlength{\itemsep}{1pt}
\item We propose GUI-PRISM, a hybrid action decoder that extends direct prediction beyond spatial grounding to complete GUI actions, supporting operation types and heterogeneous arguments through structured prediction and selective language generation.
\item We introduce field-specific queries that extract relevant multimodal information while incorporating preceding predictions to preserve dependencies among action fields.

\item We demonstrate gains in GUI grounding and complete-action prediction with lower latency, supported by ablations on field dependencies and prediction refinement.

\end{itemize}

\begin{figure*}[t]
\centering\includegraphics[width=0.99\textwidth]{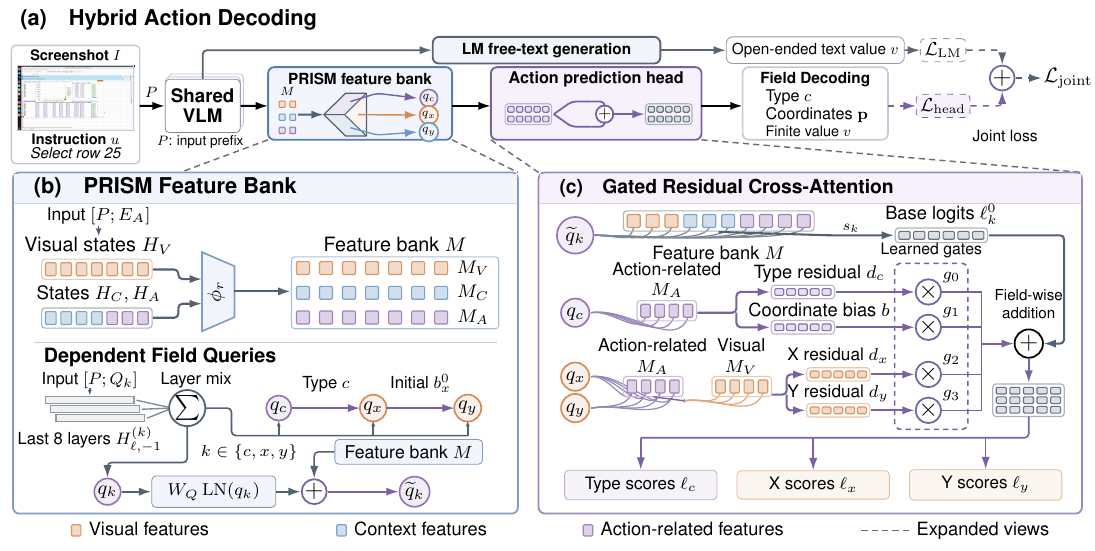}
\caption{\textbf{GUI-PRISM architecture.}
(a) Given a screenshot $I$ and instruction $u$, the shared VLM processes the input prefix $P$. The action head predicts the action type $c$, target coordinate $\mathbf p$, and finite-valued arguments, while the LM generates open-ended text values when required. These outputs form the complete action $a=(c,v,\mathbf p)$.
(b) The PRISM feature bank projects visual, contextual, and action-relevant states into shared memory $M$. Field-specific suffixes $Q_k$ produce queries that incorporate preceding predictions and memory summaries, yielding memory-conditioned queries $\widetilde q_k$.
(c) Each query reads $M$ to produce base logits. Gated residuals derived from action-relevant features $M_A$ and visual features $M_V$ refine them into final type and coordinate logits $\ell_c,\ell_x,\ell_y$ for field decoding.}
\label{fig:head}
\end{figure*}

\section{Related Work}

\par\noindent\textbf{GUI agents.}\quad
GUI agents translate visual observations and user instructions into executable actions. Early approaches develop screenshot understanding and visual grounding through multimodal pretraining and large-scale GUI data~\cite{hong2023cogagent,cheng2024seeclick,gou2024uground,wu2024osatlas}. Subsequent methods learn action prediction from interaction demonstrations, incorporating task context and intermediate reasoning to support operation selection and target localization~\cite{xu2024aguvis,qin2025uitars,wang2025opencua,liu2026benchmarking}. Reinforcement learning further improves agent behavior through feedback on action correctness and task outcomes~\cite{yang2026guilibra,yang2025gta1,gu2025uivenus}. These advances primarily improve the capabilities and training of GUI policies, while many approaches still express actions as autoregressively generated text. GUI-PRISM focuses on the action decoder, directly predicting structured fields and retaining language generation for open-ended text values.

\par\noindent\textbf{GUI grounding.}\quad
Recent methods introduce specialized mechanisms to improve target localization beyond coordinate-string generation. Attention-based approaches associate action representations with visual patches and refine spatial attention through background suppression and center-focused supervision~\cite{wu2025guiactor,chen2026v2p}. Other methods improve efficiency through position-preserving visual token selection or predict locations through continuous coordinate decoding~\cite{ouyang2026focusui,jinggrounding}. These approaches primarily address the spatial component of GUI interaction. Complete action prediction additionally requires selecting operation types and producing their associated arguments, including discrete values and free text. GUI-PRISM extends direct prediction to these heterogeneous fields and explicitly uses preceding predictions to guide subsequent ones.

\par\noindent\textbf{Structured prediction.}\quad
Query-based decoders provide a general mechanism for extracting output-specific information from shared features. DETR uses object queries for detection, while Perceiver IO uses output queries to support structured outputs~\cite{carion2020detr,jaegle2022perceiverio}. For spatial prediction, TokenPose represents keypoints with learned tokens, and SimCC formulates localization as horizontal and vertical coordinate classification~\cite{li2021tokenpose,li2022simcc}. In contrast, Pix2Seq represents object categories and coordinates as autoregressive token sequences~\cite{chen2022pix2seq}. GUI-PRISM combines direct field prediction with explicit conditioning between fields: the predicted operation guides coordinate queries, and the initial horizontal-coordinate estimate informs vertical-coordinate prediction. Field-specific queries retrieve relevant evidence from shared multimodal memory, while gated residual branches refine the resulting predictions.

\section{Method}
\label{sec:method}

\subsection{Problem Formulation}
\label{sec:problem-formulation}

Given a screenshot $I$ and task context $u$, including the user instruction and available interaction history, our goal is to predict the next executable GUI action $a=(c,v,\mathbf p)$. Here $c\in\mathcal C$ denotes one of 13 supported operation types, $v$ is the operation-specific value, and $\mathbf p$ is an optional target coordinate. For actions with a target coordinate, $\mathbf p=(x,y)$ contains coordinates normalized to $[0,1000]$; otherwise, $\mathbf p=\varnothing$. For example, selecting an option can be represented as $a=(\texttt{Select},\texttt{English},(420,360))$, while navigating back is represented as $a=(\texttt{NavigateBack},\texttt{None},\varnothing)$. Each operation specifies its valid arguments and target-coordinate requirements, forming the structured action space $\mathcal A$. Table~\ref{tab:action-space} summarizes the supported operations and their required fields.
Formally, GUI-PRISM predicts the action as
\begin{equation}
a=f_\theta(I,u)\in\mathcal A,
\label{eq:action-mapping}
\end{equation}
where $\theta$ denotes the model parameters. As illustrated in Fig.~\ref{fig:head}, GUI-PRISM combines direct prediction of structured fields with language generation for open-ended text values, while modeling dependencies among action fields.

\subsection{Hybrid Action Decoding}
\label{sec:structured-decoding}

GUI-PRISM constructs a shared feature bank and employs dependent field queries to predict complete GUI actions, allowing preceding predictions to guide subsequent fields.

\par\noindent\textbf{PRISM Feature Bank.}\quad
Accurate GUI action prediction requires complementary visual, contextual, and action-relevant information. To make these information sources available for different action fields, we construct a shared PRISM feature bank that integrates their representations. Let $P$ denote the input prefix containing the observable multimodal input and inherited learned context positions. We introduce six learnable embeddings $E_A$ to capture complementary action-relevant information for prediction.
Processing $[P;E_A]$ through the VLM backbone yields contextual states $H_C$ from the inherited context positions, action-relevant states $H_A$ from the additional embeddings, and visual states $H_V$ from image-token positions.

Each feature group is mapped to $M_r=\phi_r(H_r)$ for $r\in\{C,A,V\}$, where $\phi_r$ incorporates layer normalization (LN), feature projection, and feature-type embeddings. The resulting feature bank is
\begin{equation}
M=[M_C;M_A;M_V]\in\mathbb R^{N\times d},\quad d=1024,
\end{equation}
where $N$ denotes the total number of feature positions. The feature bank is computed once per input and shared across field predictions, with padding masked.

\par\noindent\textbf{Dependent Field Queries.}\quad
While the shared PRISM feature bank provides multimodal information, different action fields require distinct information and exhibit dependencies during prediction. To address these requirements, we construct three dependent field prompts $Q_c$, $Q_x$, and $Q_y$ for operation and coordinate prediction. The coordinate prompts incorporate the selected operation, while $Q_y$ additionally includes the initial X estimate.

Each prompt is processed together with $P$, and its hidden states are aggregated through layer mixing:
\begin{equation}
q_k
=
\sum_{\ell=1}^{8}
\operatorname{softmax}(\alpha)_\ell
H_{\ell,-1}^{(k)}
\in\mathbb R^{4096},
\quad k\in\{c,x,y\},
\label{eq:mix}
\end{equation}
where $H_{\ell,-1}^{(k)}$ denotes the final-token hidden state of the corresponding field branch at layer $\ell$, and $\alpha\in\mathbb R^8$ contains shared learnable mixing logits.

To incorporate information from different feature groups, we compute a padding-masked mean feature $\mu_\rho\in\mathbb R^d$ for each available group $\rho\in\mathcal R$, where $\mathcal R\subseteq\{C,A,V\}$. Each query combines its projected representation with a learned weighted sum of these features:
\begin{equation}
\widetilde q_k
=
W_Q\operatorname{LN}(q_k)
+
\sum_{\rho\in\mathcal R}
\operatorname{softmax}_{\rho}(\gamma_{k,\rho})
\mu_\rho,
\end{equation}
where $W_Q\in\mathbb R^{d\times4096}$ projects the query and $\gamma_{k,\rho}$ controls the contribution of each feature group, with the softmax normalized over available groups.

Each query $\widetilde q_k$ attends to the PRISM feature bank $M$ through shared cross-attention, followed by a shared FFN to produce $s_k\in\mathbb R^d$. Linear classifiers yield initial operation logits $\ell_c^0\in\mathbb R^{13}$ and coordinate-bin logits $\ell_x^0,\ell_y^0\in\mathbb R^{512}$, which are subsequently refined by gated residual branches.

\begin{table*}[t]
\centering
\caption{Grounding accuracy (\%) and inference cost on ScreenSpot-v2 under native protocols. GUI-Actor, V2P, InfiGUI-G1, and UI-Venus-Ground use published accuracy and separately measured local latency and token counts. Bold and underlined scores indicate the best and second-best results \mbox{within each group}.}
\label{tab:ssv2-main}

\small
\setlength{\tabcolsep}{2pt}
\renewcommand{\arraystretch}{1.08}

\begin{tabular*}{\textwidth}{@{\extracolsep{\fill}}lrrrrrrrrr@{}}
\toprule
\multirow{2}{*}{Method / configuration} & \multicolumn{2}{c}{Mobile} & \multicolumn{2}{c}{Desktop} & \multicolumn{2}{c}{Web} & \multirow{2}{*}{Overall $\uparrow$} & \multirow{2}{*}{Tokens $\downarrow$} & \multirow{2}{*}{Time (s) $\downarrow$}\\
\cmidrule(lr){2-3}\cmidrule(lr){4-5}\cmidrule(lr){6-7}
& Text & Icon & Text & Icon & Text & Icon & & &\\
\midrule
\rowcolor{black!7}
\multicolumn{10}{@{}l}{Autoregressive coordinate generation}\\
OpenCUA-7B~\cite{wang2025opencua} & 93.8 & 88.6 & 96.9 & 87.1 & 88.0 & 83.3 & 89.9 & 23.3 & 1.7\\
GTA1-7B~\cite{yang2025gta1} & \underline{99.0} & 88.6 & 94.9 & 89.3 & 92.3 & 86.7 & 92.4 & \underline{12.4} & \textbf{1.2}\\
GUI-G$^2$-7B~\cite{tang2025guig2} & 98.3 & \underline{91.9} & 95.4 & 89.3 & 94.0 & 87.7 & 93.3 & 18.8 & 1.5\\
InfiGUI-G1-7B (Top-1)~\cite{liu2025infiguig1} & \underline{99.0} & \underline{91.9} & 94.3 & 82.1 & \underline{97.9} & 89.2 & 93.5 & 93.8 & 4.3\\
UI-Venus-Ground-7B~\cite{gu2025uivenus} & \underline{99.0} & 90.0 & 97.0 & \underline{90.7} & 96.2 & 88.7 & 94.1 & 19.0 & 1.5\\
POINTS-GUI-G-8B~\cite{zhao2026pointsguig} & \underline{99.0} & 90.5 & \textbf{99.5} & \textbf{92.9} & 94.9 & \underline{92.6} & \underline{95.2} & 14.7 & 2.0\\
UI-Venus-1.5-8B~\cite{gao2026uivenus} & \textbf{99.3} & \textbf{92.9} & 96.4 & \textbf{92.9} & \textbf{98.3} & \textbf{93.1} & \textbf{95.9} & \textbf{10.7} & \underline{1.4}\\
Qwen3-VL-8B-Instruct~\cite{bai2025qwen3vl} & 94.1 & 89.6 & \underline{98.5} & 90.0 & 94.9 & 88.2 & 92.8 & 45.5 & 3.1\\
GUI-Libra-8B (ASFT+RL, thinking)~\cite{yang2026guilibra} & 97.6 & 87.7 & 95.9 & 85.0 & 91.5 & 81.8 & 90.6 & 202.6 & 12.3\\
GUI-Owl-1.5-8B~\cite{xu2026mobileagent35} & 98.6 & 90.5 & 97.9 & 87.1 & 95.7 & 87.7 & 93.6 & 34.7 & 2.8\\
Qwen3-VL-8B~\cite{bai2025qwen3vl} (action-only SFT) & 92.8 & 87.7 & 95.9 & 90.0 & 93.6 & 89.2 & 91.7 & 45.4 & --\\
MAI-UI-8B (local)~\cite{zhou2025maiui} & \textbf{99.3} & 88.6 & \underline{98.5} & 90.0 & \underline{97.9} & 90.6 & 94.7 & 80.4 & 5.3\\
\midrule[0.8pt]
\rowcolor{black!7}
\multicolumn{10}{@{}l}{Head-based spatial decoding}\\
GGA-8B (published)~\cite{jinggrounding} & 98.5 & \underline{90.6} & 96.1 & 85.5 & 94.8 & 86.6 & 92.0 & -- & --\\
GUI-Actor-7B~\cite{wu2025guiactor} & 97.6 & 88.2 & \underline{96.9} & 85.7 & 93.2 & \underline{86.7} & 92.1 & \underline{1.0} & \textbf{0.8}\\
FocusUI-7B ($r=100\%$)~\cite{ouyang2026focusui} & \underline{98.8} & \textbf{91.6} & 95.6 & \underline{92.1} & 95.0 & 84.4 & \underline{93.1} & -- & --\\
V2P-7B~\cite{chen2026v2p} & 98.1 & 88.0 & 96.1 & 89.7 & \underline{95.4} & 84.4 & 92.4 & \underline{1.0} & \textbf{0.8}\\
\textbf{GUI-PRISM (Ours)} & \textbf{99.0} & 89.6 & \textbf{98.5} & \textbf{92.9} & \textbf{97.9} & \textbf{90.1} & \textbf{95.1} & \textbf{0.0} & \textbf{0.8}\\
\bottomrule
\end{tabular*}
\end{table*}

\par\noindent\textbf{Gated Residual Cross-Attention.}\quad
While field-specific queries capture relevant multimodal information for initial predictions, fine-grained action-relevant and visual information is essential for accurate operation selection and coordinate localization. To further incorporate such information, we introduce gated residual cross-attention to refine the initial predictions. Specifically, the operation branch attends to action-relevant features $M_A$, while the coordinate branches attend to both $M_A$ and visual features $M_V$, producing corrections $d_c\in\mathbb R^{13}$ and $d_x,d_y\in\mathbb R^{512}$.

We additionally predict coordinate-permission logits $r\in\mathbb R^2$. Let $G(c)\in\{0,1\}$ indicate whether operation $c$ permits coordinates.
Let $\pi=\log\operatorname{softmax}(r)$. The operation bias is
\begin{equation}
b_c=\pi_{G(c)}-\frac{1}{|\mathcal C|}
\sum_{j\in\mathcal C}\pi_{G(j)},
\end{equation}
where $b\in\mathbb R^{13}$ collects the operation biases.
Four learnable gates $g_i=\sigma(\eta_i)$, $i\in\{0,1,2,3\}$, control the residual contributions:
\begin{equation}
\begin{aligned}
\ell_c &= \ell_c^0+g_0d_c+g_1b,\\
\ell_x &= \ell_x^0+g_2d_x,\quad
\ell_y=\ell_y^0+g_3d_y.
\end{aligned}
\end{equation}
Here, the refined logits $\ell_c$, $\ell_x$, and $\ell_y$ are used for operation and coordinate prediction.

\par\noindent\textbf{Field Decoding.}\quad
We predict structured action fields in dependency order without token-by-token generation. The predicted operation $c=\arg\max_{j\in\mathcal C}(\ell_c)_j$ determines the applicable arguments and guides coordinate prediction. The Y query uses the initial X estimate $b_x^0=\arg\max_b\ell_{x,b}^0$, mapped to $[0,1000]$. For each axis $z\in\{x,y\}$, we select $b_z=\arg\max_b\ell_{z,b}$ and obtain $p_z=\operatorname{round}(1000b_z/511,3)$.

Finite-valued arguments are predicted by operation-specific classifiers applied to $s_c$, while open-ended text values are generated by the language model. Actions requiring coordinates always include them, whereas actions with optional coordinates use a separate coordinate-presence prediction. The action value is
\begin{equation}
v =
\begin{cases}
h_v(s_c,c), & c\in\mathcal C_{\rm finite},\\
\operatorname{LM}_{\rm value}(P,c), & c\in\mathcal C_{\rm text},\\
\texttt{None}, & \text{otherwise},
\end{cases}
\label{eq:value-decoding}
\end{equation}
where $\mathcal C_{\rm finite}$ and $\mathcal C_{\rm text}$ denote operations requiring finite-valued and open-ended text arguments. Finally, the predicted fields are assembled into an executable action $a=\operatorname{Serialize}(c,v,\mathbf p)$ according to Table~\ref{tab:action-space}.

\begin{table*}[t]
\centering
\caption{Complete-action accuracy (\%) on AndroidControl-v2 and MM-Mind2Web-v2. AC Overall and MM Overall denote the equally weighted means across two instruction levels and three evaluation splits, respectively. Bold and underlined scores indicate the best and second-best results, respectively.}
\label{tab:head-nav}
\small
\setlength{\tabcolsep}{3pt}
\renewcommand{\arraystretch}{1.08}
\begin{tabular*}{\textwidth}{@{\extracolsep{\fill}}lrrrrrrr@{}}
\toprule
\multirow{2}{*}{Model / configuration} & \multicolumn{3}{c}{AndroidControl-v2} & \multicolumn{4}{c}{MM-Mind2Web-v2}\\
\cmidrule(lr){2-4}\cmidrule(lr){5-8}
 & High & Low & Overall $\uparrow$ & Cross-task & Cross-website & Cross-domain & Overall $\uparrow$\\
\midrule
\rowcolor{black!7}
\multicolumn{8}{@{}l}{Native-model references}\\
GUI-R1-3B~\cite{luo2025guir1} & 40.0 & 55.8 & 47.9 & 24.0 & 22.3 & 24.6 & 23.6\\
GUI-R1-7B~\cite{luo2025guir1} & 39.7 & 62.3 & 51.0 & 37.0 & 34.1 & 39.6 & 36.9\\
Aguvis-7B~\cite{xu2024aguvis} & 37.7 & 48.0 & 42.9 & 37.7 & 31.7 & 36.9 & 35.4\\
UI-TARS-1.5-7B~\cite{qin2025uitars} & 45.2 & 48.5 & 46.9 & 37.2 & 31.3 & 35.6 & 34.7\\
GLM-4.1V-9B-Thinking~\cite{glmteam2025glm4v} & 37.2 & 67.1 & 52.2 & 26.9 & 23.0 & 28.7 & 26.2\\
Qwen2.5-VL-3B~\cite{bai2025qwen25vl} & 36.4 & 71.1 & 53.8 & 24.4 & 18.6 & 27.1 & 23.4\\
Qwen2.5-VL-7B~\cite{bai2025qwen25vl} & 46.5 & 67.8 & 57.2 & 31.6 & 30.4 & 35.6 & 32.5\\
Qwen3-VL-4B~\cite{bai2025qwen3vl} & 49.3 & \underline{78.9} & 64.1 & 42.9 & 38.8 & 42.0 & 41.2\\
Qwen3-VL-8B~\cite{bai2025qwen3vl} & 54.8 & 77.6 & 66.2 & 44.7 & 41.0 & 45.6 & \underline{43.8}\\
GUI-Libra-8B~\cite{yang2026guilibra} & 56.8 & 70.1 & 63.4 & \underline{45.0} & \underline{41.5} & 44.2 & 43.6\\
\midrule
\rowcolor{black!7}
\multicolumn{8}{@{}l}{Multi-module frameworks}\\
GPT-4o + UGround-v1-7B~\cite{yang2026guilibra} & 57.0 & 78.4 & 67.7 & 35.7 & 33.9 & 39.1 & 36.2\\
GPT-4.1 + UGround-v1-7B~\cite{yang2026guilibra} & \underline{57.5} & 78.4 & \underline{68.0} & 41.1 & 36.2 & 43.0 & 40.1\\
GPT-5-mini + UGround-v1-7B~\cite{yang2026guilibra} & 52.8 & 77.1 & 65.0 & 44.2 & 40.4 & \underline{45.8} & 43.5\\
\midrule
\rowcolor{black!7}
\multicolumn{8}{@{}l}{Local hybrid action decoding}\\
\textbf{GUI-PRISM (Ours)} & \textbf{58.3} & \textbf{83.7} & \textbf{71.0} & \textbf{45.7} & \textbf{41.6} & \textbf{46.1} & \textbf{44.5}\\
\bottomrule
\end{tabular*}
\end{table*}

\subsection{Training}
\label{sec:training-inference}

To jointly optimize structured action prediction and open-ended text generation, we train GUI-PRISM with complementary supervision objectives.

\par\noindent\textbf{Action prediction supervision.}\quad
We use cross-entropy losses $\mathcal L_{\rm type}$, $\mathcal L_x$, and $\mathcal L_y$ to supervise operation types and coordinate bins. Additional cross-entropy losses supervise coordinate presence ($\mathcal L_{\rm pres}$), finite-valued arguments ($\mathcal L_{\rm finite}$), and coordinate permission ($\mathcal L_{\rm pointer}$). The auxiliary loss is
\begin{equation}
\mathcal L_{\rm aux}=\lambda_p\mathcal L_{\rm pres}+\mathcal L_{\rm finite}+\mathcal L_{\rm pointer},
\end{equation}
where $\lambda_p=0.25$. The action-head objective is
\begin{equation}
\mathcal L_{\rm head}=\mathcal L_{\rm type}+m(\mathcal L_x+\mathcal L_y)+\mathcal L_{\rm aux},
\end{equation}
where $m\in\{0,1\}$ indicates whether the ground-truth action contains a target coordinate.

\par\noindent\textbf{Joint optimization.}\quad
The main training objective combines action prediction and language generation:
\begin{equation}
\mathcal L_{\rm joint}=\lambda_{\rm h}\mathcal L_{\rm head}+\lambda_{\rm L}\mathcal L_{\rm LM},
\end{equation}
where $\lambda_{\rm h}=\lambda_{\rm L}=1$ are the loss weights, and $\mathcal L_{\rm LM}$ supervises native action serialization.

\section{Experiments}
\label{sec:experiments}

We evaluate GUI-PRISM on GUI grounding and offline next-action prediction benchmarks, examining both prediction accuracy and inference efficiency. We further analyze its decoder design through action-head comparisons, field-level evaluations, and component ablations.

\subsection{Experimental Setup}

\par\noindent\textbf{Benchmarks and Metrics.}\quad
We evaluate GUI grounding on OSWorld-G~\cite{xie2025osworldg}, ScreenSpot-v2 (SS-v2), and the English subset of ScreenSpot-Pro (SS-Pro)~\cite{li2025screenspotpro}, containing 564, 1,272, and 1,581 examples, respectively. Grounding accuracy is measured against the original target geometry, with parsing failures counted as errors.

For offline next-action prediction, we use the GUI-Libra adaptations~\cite{yang2026guilibra} of AndroidControl~\cite{li2024androidcontrol} and Mind2Web~\cite{deng2023mind2web}. AndroidControl-v2 (AC-v2) contains 398 examples at each instruction level, while MM-Mind2Web-v2 contains 1,328, 1,019, and 1,002 examples across its Cross-Task, Cross-Website, and Cross-Domain splits. Overall accuracy equally averages the corresponding levels or splits. Field-level evaluation uses all SS-v2 examples; for AC-v2, coordinate and joint accuracies are evaluated on the 223 examples with target coordinates, while action-type accuracy uses all examples at each level.

\par\noindent\textbf{Training Configuration.}\quad
GUI-PRISM is initialized from Qwen3-VL-8B and trained through staged adaptation. We first train the action decoder on 81,418 GUI-Libra-derived examples~\cite{yang2026guilibra}, followed by joint adaptation of the decoder and backbone. Additional grounding supervision is incorporated from GroundCUA, Aguvis, and OS-Atlas~\cite{feizi2025groundcua,xu2024aguvis,wu2024osatlas}. Training uses eight PPU-ZW810E accelerators with 96\,GiB of memory each, a global batch size of 32, and a maximum image budget of 2,109,744 pixels. More training details are provided in Supplementary Material.

\par\noindent\textbf{Evaluation Protocol.}\quad
Local inference uses batch size 1 and image budgets ranging from 4,096 to 16,777,216 pixels unless otherwise specified. GUI-PRISM directly predicts operation types and their associated structured arguments without forced action, coordinate repair, or test-time search.

\begin{figure*}[t]
\centering
\includegraphics[width=\textwidth]{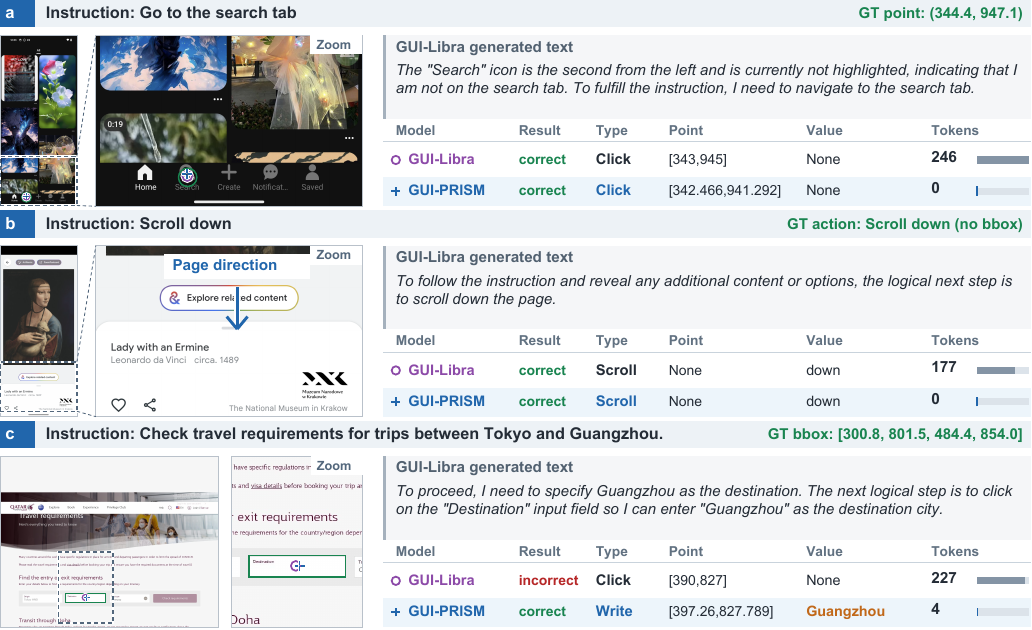}
\caption{\textbf{Complete action predictions.} GUI-PRISM and GUI-Libra (ASFT + RL) on AC-v2 Low (a,b) and  MM-v2 Cross-Task (c). Italics quote GUI-Libra; token counts include reasoning and delimiters. Rows show action type, coordinate, and value. \textcolor[HTML]{2166AC}{Blue crosses}: GUI-PRISM; \textcolor[HTML]{8C4AA8}{Purple circles}: GUI-Libra; \textcolor[HTML]{198450}{Green}: targets; \textcolor[HTML]{C16819}{Orange}: generated text. Dashed boxes mark zoom regions; (b): arrow shows page direction. GT labels retain the original coordinate (a), action (b), or bbox (c). Coordinates use $[0,1000]$; bbox entries are $[x_{\min},y_{\min},x_{\max},y_{\max}]$.}
\label{fig:complete-action-cases}
\end{figure*}

\subsection{GUI Grounding}

\par\noindent\textbf{Accuracy.}\quad
As shown in Table~\ref{tab:ssv2-main}, GUI-PRISM achieves 95.1\% accuracy on SS-v2, outperforming all compared head-based grounding methods. Specifically, it exceeds FocusUI, V2P, GUI-Actor, and GGA by 2.0, 2.7, 3.0, and 3.1 percentage points, respectively. 
These results highlight the effectiveness of GUI-PRISM in spatial grounding while supporting complete action prediction.

Across platform and element categories, GUI-PRISM achieves 99.0\% accuracy on mobile text, 98.5\% on desktop text, and 97.9\% on web text. It also reaches 92.9\% on desktop icons, matching the highest score among the listed methods. Compared with FocusUI, the strongest head-based baseline in overall accuracy, GUI-PRISM improves desktop and web text localization by 2.9 points each, and web icon localization by 5.7 points, although FocusUI performs better on mobile icons. These results indicate that GUI-PRISM improves grounding across different interface settings rather than relying on a single category.

Compared with autoregressive approaches, GUI-PRISM improves upon Qwen3-VL-8B-Instruct by 2.3 percentage points and exceeds the locally evaluated MAI-UI-8B by 0.4 points. Its overall accuracy remains competitive with POINTS-GUI-G-8B and UI-Venus-1.5-8B.

\par\noindent\textbf{Inference Efficiency.}\quad
\label{sec:inference-cost}
GUI-PRISM requires zero autoregressively generated output tokens for grounding, as operation types and coordinates are predicted directly. In contrast, autoregressive methods generate 10.7 to 202.6 tokens, from UI-Venus-1.5-8B to GUI-Libra, as reported in Table~\ref{tab:ssv2-main}.

GUI-PRISM achieves 95.1\% accuracy on SS-v2 with an average inference time of 0.8 seconds per example, compared with 3.1 seconds for Qwen3-VL-8B-Instruct, 5.3 seconds for MAI-UI-8B, and 2.0 seconds for POINTS-GUI-G-8B. It also matches the inference latency of GUI-Actor and V2P while achieving higher grounding accuracy.

\renewcommand{\topfraction}{0.6}
\begin{table}[!t]
\centering\small
\caption{Grounding accuracy (\%) of different action heads on OSWorld-G and SS-v2. Avg. denotes the equally weighted mean. 
}
\label{tab:action-head-comparison}
\begin{tabular*}{\columnwidth}{@{\extracolsep{\fill}}lrrr@{}}
\toprule
Action head & OS-G & SS-v2 & Avg.\\
\midrule
Diffusion~\cite{ho2020ddpm} & 1.6 & 11.4 & 6.5\\
Flow Matching~\cite{lipman2023flowmatching} & 6.6 & 32.5 & 19.5\\
MLP & \underline{9.4} & \underline{43.6} & \underline{26.5}\\
\midrule
GUI-PRISM head & \textbf{33.5} & \textbf{85.2} & \textbf{59.4}\\
\bottomrule
\end{tabular*}
\end{table}

\subsection{Offline Action Prediction}

\par\noindent\textbf{Accuracy.}\quad
As shown in Table~\ref{tab:head-nav}, we evaluate complete GUI actions on AC-v2 and MM-v2 under GUI-Libra protocols~\cite{yang2026guilibra}. Unlike spatial grounding, these benchmarks evaluate both operation types and their arguments. GUI-PRISM achieves 71.0\% accuracy on AC-v2 and 44.5\% on MM-v2, exceeding the reported Qwen3-VL-8B baseline by 4.8 and 0.7 percentage points, respectively. It also outperforms the other methods listed in the table on both overall metrics.

On AC-v2, GUI-PRISM achieves 58.3\% and 83.7\% accuracy under High and Low instruction levels, improving upon Qwen3-VL-8B by 3.5 and 6.1 percentage points, respectively. On MM-v2, it achieves 45.7\%, 41.6\%, and 46.1\% accuracy across the Cross-Task, Cross-Website, and Cross-Domain splits, yielding consistent improvements of 1.0, 0.6, and 0.5 percentage points.

\par\noindent\textbf{Inference Efficiency.}\quad
We further evaluate complete-action inference efficiency using Qwen3-VL-8B with direct autoregressive action generation as the baseline. Under the evaluated deployment configurations, GUI-PRISM reduces average latency from 2.55 to 0.93 seconds on AC-v2 and from 2.30 to 0.60 seconds on MM-v2, achieving speedups of 2.7$\times$ and 3.9$\times$, respectively. Timing includes visual encoding, action prediction, and any required text generation.

Action types, coordinates, and scroll directions are predicted directly, while the language model generates open-ended text values. GUI-PRISM generates an average of 0.42 and 0.70 tokens per action on AC-v2 and MM-v2, respectively, compared with 31.06 and 33.04 for Qwen3-VL-8B. These averages count language-model-generated tokens across all examples, including zero-token predictions.

\subsection{Qualitative Analysis}
\label{sec:complete-action-cases}
Fig.~\ref{fig:complete-action-cases} compares GUI-Libra (ASFT + RL) and the GUI-PRISM model. Both select the search tab in Fig.~\ref{fig:complete-action-cases}(a) and predict \texttt{Scroll} with value \texttt{down} and no coordinate in Fig.~\ref{fig:complete-action-cases}(b). GUI-PRISM uses zero autoregressive output tokens, versus 246 and 177 for GUI-Libra, demonstrating unified decoding of spatial and direction-valued actions.

In Fig.~\ref{fig:complete-action-cases}(c), GUI-Libra states ``\texttt{specify Guangzhou as the destination}'' but outputs \texttt{Click} with no text value. Both coordinates fall within the GT box, so the decisive difference is the operation and value, not localization. GUI-PRISM returns \texttt{Write} with \texttt{Guangzhou}, matching the annotated next action with four generated tokens versus 227. These native-configuration examples illustrate complete-action correctness; token counts alone do not measure end-to-end speedup.

\setcounter{topnumber}{2}
\begin{table}[t]
\centering\footnotesize
\setlength{\tabcolsep}{3pt}
\renewcommand{\arraystretch}{1.06}
\caption{Field-level accuracy (\%) on SS-v2 and AC-v2.
X and Y measure horizontal and vertical localization accuracy, respectively; XY requires a coordinate inside the target box. T measures action-type accuracy, while J requires both T and XY to be correct.}
\label{tab:field-accuracy}
\begin{tabular*}{\columnwidth}{@{\extracolsep{\fill}}lrrrrr@{}}
\toprule
\rowcolor{black!7}
\multicolumn{6}{@{}l}{SS-v2}\\
\midrule
\multicolumn{2}{@{}l}{Model} & X & \multicolumn{2}{c}{Y} & XY\\
\midrule
\multicolumn{2}{@{}l}{Qwen3-VL~\cite{bai2025qwen3vl}}
& 93.9 & \multicolumn{2}{c}{95.4} & 92.8\\
\multicolumn{2}{@{}l}{GUI-Owl~\cite{xu2026mobileagent35}}
& \underline{95.7} & \multicolumn{2}{c}{\underline{96.1}}
& \underline{93.6}\\
\multicolumn{2}{@{}l}{\textbf{GUI-PRISM}}
& \textbf{96.8} & \multicolumn{2}{c}{\textbf{97.4}}
& \textbf{95.1}\\
\midrule
\rowcolor{black!7}
\multicolumn{6}{@{}l}{AndroidControl-v2}\\
\midrule
Model & T & X & Y & XY & J\\
\midrule
\multicolumn{6}{@{}l}{\textit{High-level Instructions}}\\
\midrule
Qwen3-VL~\cite{bai2025qwen3vl}
& \underline{56.3} & \underline{64.1} & \underline{61.0}
& \underline{57.0} & \underline{55.6}\\
GUI-Owl~\cite{xu2026mobileagent35}
& 21.4 & 51.6 & 52.0 & 44.4 & 16.6\\
\textbf{GUI-PRISM}
& \textbf{70.4} & \textbf{77.1} & \textbf{78.9}
& \textbf{72.6} & \textbf{72.6}\\
\midrule
\multicolumn{6}{@{}l}{\textit{Low-level Instructions}}\\
\midrule
Qwen3-VL~\cite{bai2025qwen3vl}
& \textbf{90.5} & 74.9 & 73.1 & 71.3 & \underline{71.3}\\
GUI-Owl~\cite{xu2026mobileagent35}
& 59.8 & \underline{87.9} & \underline{83.0}
& \underline{79.8} & 63.2\\
\textbf{GUI-PRISM}
& \underline{88.9} & \textbf{91.5} & \textbf{91.0}
& \textbf{89.2} & \textbf{89.2}\\
\bottomrule
\end{tabular*}
\end{table}

\subsection{Ablation and Analysis}

\par\noindent\textbf{Choice of Action Head.}\quad
Table~\ref{tab:action-head-comparison} compares different action heads under matched training settings, using 81,418 examples with the backbone frozen. GUI-PRISM achieves the highest mean accuracy of 59.4\%, substantially outperforming MLP at 26.5\%, Flow Matching at 19.5\%, and Diffusion at 6.5\%. Unlike trajectory-based action generation, this evaluation focuses on single-step GUI grounding. GUI-PRISM directly predicts coordinate bins while leveraging field-specific multimodal information and coordinate dependencies. These results demonstrate the effectiveness of its structured decoding design, although differences in training objectives and model architectures prevent attributing the gains to any single component.

\par\noindent\textbf{Field-level Prediction.}\quad
Table~\ref{tab:field-accuracy} evaluates operation selection and coordinate localization at the field level. On SS-v2, GUI-PRISM achieves 96.8\% and 97.4\% accuracy on the X and Y axes, respectively, reaching 95.1\% joint localization accuracy, 1.5 percentage points above GUI-Owl. On AC-v2 High, GUI-PRISM improves operation-type accuracy from 56.3\% to 70.4\% and joint action accuracy from 55.6\% to 72.6\% over Qwen3-VL. On AC-v2 Low, despite a 1.6-point decrease in operation-type accuracy, both joint localization and joint action accuracy improve by 17.9 points, reaching 89.2\%. These results highlight consistent gains in spatial grounding, while improvements in operation selection vary across instruction levels.

\setcounter{topnumber}{2}
\begin{table}[t]
\centering\footnotesize
\caption{Component ablations on AC-v2.
FQ: field queries; FD: field dependencies;
GR: gated residual module.
Avg. equally weights High and Low accuracy (\%).
Bold and underlined scores indicate the best and second-best results, respectively.}
\label{tab:component-ablation}
\setlength{\tabcolsep}{3pt}
\renewcommand{\arraystretch}{1.08}
\begin{tabular*}{\columnwidth}{@{\extracolsep{\fill}}lcccrrr@{}}
\toprule
Configuration & FQ & FD & GR & High & Low & Avg.\\
\midrule
MLP
& -- & -- & --
& 11.6 & 26.6 & 19.1\\
Independent queries
& $\checkmark$ & -- & --
& 24.4 & 41.7 & 33.0\\
Dependent queries
& $\checkmark$ & $\checkmark$ & --
& \textbf{49.0} & \underline{71.6} & \underline{60.3}\\
GUI-PRISM
& $\checkmark$ & $\checkmark$ & $\checkmark$
& \underline{48.2} & \textbf{72.9} & \textbf{60.6}\\
\bottomrule
\end{tabular*}
\end{table}

\par\noindent\textbf{Component Ablations.}\quad
Table~\ref{tab:component-ablation} compares separately trained configurations on AC-v2 at High and Low instruction levels. All configurations use the same 81,418 training examples and matched head-only training settings.

The MLP baseline achieves 11.6\% and 26.6\% accuracy under High and Low instructions, respectively. Introducing independent field queries improves these scores to 24.4\% and 41.7\%, increasing mean accuracy from 19.1\% to 33.0\%. Adding operation-to-coordinate and X-to-Y dependencies further raises accuracy to 49.0\% and 71.6\%, yielding a mean of 60.3\%. Both instruction levels benefit from field queries and dependency modeling. These gains support field-specific feature extraction and explicitly passing preceding decisions to subsequent predictions. Adding the gated residual module further improves Low accuracy to 72.9\%, while High accuracy changes to 48.2\%. The complete GUI-PRISM head achieves the highest mean accuracy of 60.6\%. Together, these incremental comparisons support the effectiveness of field queries, field dependencies, and the gated residual module.

\section{Conclusion}
In this paper, we propose GUI-PRISM, a hybrid action decoder
for efficient and accurate GUI action prediction.
GUI-PRISM directly predicts action types and structured arguments
while retaining language generation for open-ended text.
It constructs a shared PRISM feature bank from visual, contextual,
and action-relevant features. Field-specific queries extract
action-relevant semantic and spatial information from this bank,
while conditional queries model dependencies among action components
by using preceding predictions to guide subsequent ones.
Gated residual cross-attention further refines these predictions
using fine-grained multimodal information.
Experiments on GUI grounding and offline action prediction
benchmarks demonstrate that GUI-PRISM improves prediction accuracy
while reducing inference latency, supporting efficient decoding
of complete GUI actions across diverse tasks and interface settings.

\FloatBarrier

{\small\bibliographystyle{ieeenat_fullname}\bibliography{references}}
\end{document}